# ASL Video Corpora & Sign Bank: Resources Available through the American Sign Language Linguistic Research Project (ASLLRP)

Carol Neidle[1], Augustine Opoku[2], and Dimitris Metaxas[3]*

1. Boston University; carol@bu.edu
2. Boston University; augustine.opoku@gmail.com
3. Rutgers University; dnm@rutgers.edu

**Abstract**

The **American Sign Language Linguistic Research Project (ASLLRP)** provides Internet access to high-quality ASL video data, generally including front and side views and a close-up of the face. The manual and non-manual components of the signing have been linguistically annotated using **SignStream**®. The recently expanded **video corpora** can be browsed and searched through the **Data Access Interface** (DAI 2) we have designed; it is possible to carry out complex searches. The data from our corpora can also be downloaded; annotations are available in an XML export format. We have also developed the **ASLLRP Sign Bank**, which contains almost 6,000 sign entries for lexical signs, with distinct English-based glosses, with a total of 41,830 examples of lexical signs (in addition to about 300 gestures, over 1,000 fingerspelled signs, and 475 classifier examples). The Sign Bank is likewise accessible and searchable on the Internet; it can also be accessed from within **SignStream**® (software to facilitate linguistic annotation and analysis of visual language data) to make annotations more accurate and efficient. Here we describe the available resources. These data have been used for many types of research in linguistics and in computer-based sign language recognition from video; examples of such research are provided in the latter part of this article.

## 1. Introduction

Access to sufficient quantities of high-quality, linguistically annotated, American Sign Language (ASL) video data from native signers, with enforcement of a 1-1 relationship between the sign production and the text-based gloss label in the annotations, is essential for research in computer-based sign language recognition from video. This article is intended to make the community aware of resources that can enable and facilitate such research, by describing the datasets that are available and the tools provided on the Web for viewing and accessing the data. The final section of this paper provides some examples of computer science research that has benefited from the shared resources.

The **Data Access Interface (DAI 2)** is a Web-based interface that has been designed to allow users to browse, search, and download linguistically annotated video data of various kinds from native signers of ASL collected as part of the **American Sign Language Linguistic Research Project** (**ASLLRP**), a collaborative venture that has included researchers from Boston University (BU), Rutgers University (RU), Rochester Institute of Technology (RIT), and Gallaudet University (GU), in collaboration also with DawnSignPress [7] (DSP).

**Overview**

There are several linguistically annotated video corpora created at Boston University that can be accessed from our website:

- *2 continuous signing corpora* [Section 2.1]

  1. the BU **ASLLRP SignStream® 3 Corpus** (2,048 utterances, 16,659 sign tokens)
  2. the older BU **NCSLGR SignStream® 2 Corpus** (1,002 utterances, 10,719 sign tokens)

- *1 corpus with isolated, citation-form signs* [Section 3.1]

  the **American Sign Language Linguistic Video Dataset (ASLLVD)** (9,674 sign tokens)

These are described below. The above continuous signing data can be **searched**, **browsed**, and **downloaded** from our **Data Access Interface (DAI 2)** < https://dai.cs.rutgers.edu/dai/s/dai >.

Our **ASLLRP Sign Bank** < https://dai.cs.rutgers.edu/dai/s/signbank > [Section 3.3], originally created to allow viewing of the citation-form **ASLLVD** data, now also makes it possible to play the corresponding segmented sign videos from our continuous signing corpora (the **ASLLRP SignStream® 3** and the **NCSLGR SignStream® 2 Corpora),** which can also be viewed in context; i.e., it is possible to play the containing video utterances.

## 2. Continuous Signing Corpora

The data available through **DAI 2** < http://dai.cs.rutgers.edu/dai/s/dai > include a large set of utterances that are part of the expanding **ASLLRP SignStream® 3 Corpus**. The video examples for most of the utterances include 3 synchronized views – front, side, and close-up of the face. **DAI 2** provides various ways to browse, search, and download these data, as described below in more detail.

### 2.1  The ASLLRP Video Corpora:  Data Sets of Continuous ASL Signing

This **ASLLRP SignStream® 3 Corpus** of is an expanding collection of continuous signing from ASL native signers, which was collected at Boston University. Synchronized videos display the signing from front and side, as well as providing a close-up view of the face. These videos were annotated with **SignStream® 3** [see Section 4] with labels and start and end frames for all events: manual English-based gloss labels, sign type, start and end handshapes (with information provided for both hands), as well as non-manual information, such as grammatical markings for things such as question status, negation, topic/focus, conditional/when and relative clauses, plus anatomical behaviors, such as head nods and shakes, eye aperture and gaze, etc.. ; See [29, 30] for annotation conventions.

As of January 2022, the BU data in this corpus consist of 43 SignStream® files containing a total of 2,048 utterances, with **16,659** total signs (representing 1,867 distinct signs or sign variants) from 4 native ASL signers.

**DawnSignPress (DSP)** recently shared 527 additional sentences from 15 native signers, with a total of 3,032 sign tokens; the **DSP** videos each have a front and side view. These can be viewed but not downloaded

This is in addition to an older dataset, available separately from the same website: the **National Center for Sign Language & Gesture Resources (NCSLGR) SignStream® 2 Corpus**, although that does not include handshape annotations. That dataset, searchable and downloadable, contains 1,002 utterances and **10,719** signed examples, annotated with glossing conventions consistent with those used for all other data available from this site; see [29, 30].

## 2.2 Access to the Continuous Signing Corpora via DAI 2

### 2.2.1 Sign-level searches

The search interface is here: < **http://dai.cs.rutgers.edu/dai/s/dai** >. Searches can be performed and constrained based on the following:

- A specific character string in the gloss label on the dominant or non-dominant hand (or either);
- Signs of particular types (lexical, number, loan, fingerspelled, gesture, classifier of various kinds, etc.); compounds, non-compounds, or either;
- Handshapes on the dominant and/or non-dominant hand, at the start and/or end; presence of reduplication or passive base arm;
- Specific **SignStream**® collection (=filename); available filenames are listed, with the number of signs contained in each listed in parentheses;
- Specific signer; available signers are listed, with the number of signs from each in parentheses.

Results are displayed in a table, as shown in **Figure 1**, which presents first the results from a search for the partial string "FATHER". In addition to the canonical form of the sign, there is also a version with reduplication of the movement (FATHER+), a version with finger wiggling (FATHERwg), and a compound meaning 'parents' composed of the sign for FATHER followed by the sign for MOTHER. The number of occurrences of each, as produced by each of the 3 participants listed, is shown in the corresponding cells, with links to the video examples. For example, there are 10 examples of the canonical form, FATHER, produced by Rachel, shown when the user clicks on the circled cell; the first two of those are displayed in this figure.

**Figure 1.** Search results for partial string "FATHER"

The SignStream® filename plus the utterance number within that collection (=file) are included in the first column of **Figure 1**. The checkboxes make it possible to add specific examples to the Download Cart. Clicking on the icons in column 4 highlighted by the dark blue oval will play the sign or entire utterance videos: front, face, or side view; an example of the facial close-up for the sign video is shown in **Figure 2**. The arrow keys make it possible to step through the video frame by frame. The filename and time within the whole video are also displayed. Clicking on the blue Gloss button visible in circled region in the bottom screen shot of **Figure 1** reveals a more complete display of the non-manual events (a rough approximation of the precise time-alignment that would be visible in the SignStream® display), also shown in Figure 2.

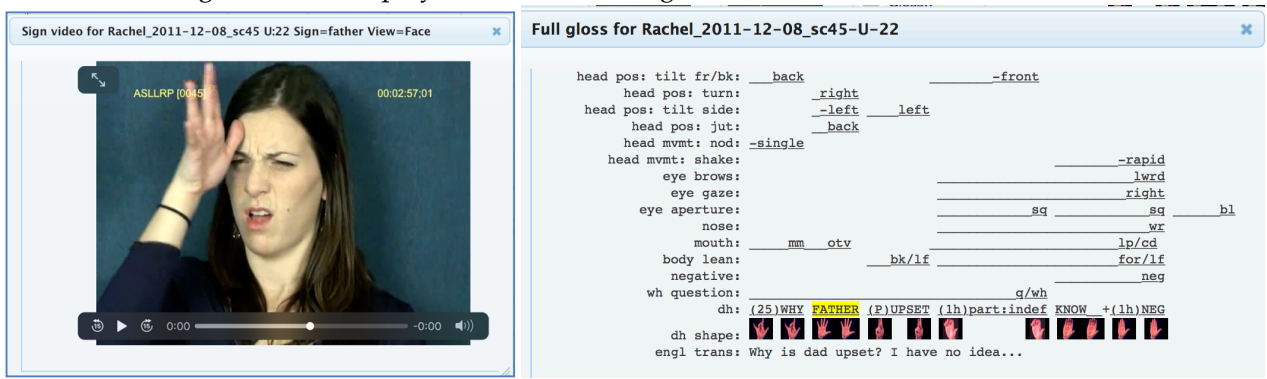

**Figure 2. Left:** Playing close-up face video for selected sign from the search results; **Right**: Display of gloss summary

### 2.2.2 Utterance-level searches

Utterance-level searches make it possible to view all utterances in the collections, or to limit the selection based on any or all of the following:

- The presence of a specific grammatical marking and/or specific anatomical non-manual actions (e.g., wh-question, negation, raised eyebrows, head nod);
- The presence of a particular character string in the English translation of the sentence;
- Specific SignStream® collection (=filename); available filenames are listed, with the number of utterances contained in each listed in parentheses;
- Specific signer; available signers are listed, with the number of utterances from each contained in parentheses.

For example, a search for the presence of negative marking would display the 532 utterances with that marking, the first two of which are shown in **Figure 3**. Each utterance video—front, face close-up, or side view—can be played.

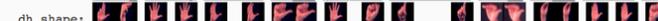

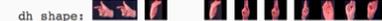
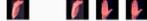

**Figure 3.** First two results of a search for utterances with negative

# 3. Isolated Signs

## 3.1 The American Sign Language Lexicon Video Dataset

The **American Sign Language Lexicon Video Dataset (ASLLVD)** is a collection of isolated, citation-form signs collected at BU through a collaborative effort of Stan Sclaroff, Vassilis Athitsos, Ashwin Thangali, and Carol Neidle, as well as many ASL consultants and BU students < **http://www.bu.edu/asllrp/people.html** >. It consists of videos of >3,300 ASL signs in citation form, each produced one or more times by each of 1-6 native ASL signers, for a total of almost 9,674 tokens [37, 46]. This dataset includes multiple synchronized videos showing the signing from different angles. Linguistic annotations [29, 30].include gloss labels, sign start and end time codes, start and end handshape labels for both hands, and morphological and articulatory classifications of sign type. For compound signs, the dataset also includes annotations for each morpheme. Further information about this data set is available from < **http://www.bu.edu/ asllrp/av/dai-asllvd.html** >.

## 3.2 Additional Datasets with Citation-Form Signs

The **ASLLRP Sign Bank** was initially constructed to allow viewing of the **ASLLVD**. It has since been expanded to incorporate display of citation-form sign examples provided by:

- the **Rochester Institute of Technology** (**RIT),** collected under the direction of Matt Huenerfauth; a total of 10,561 sign tokens
- **DawnSignPress (DSP);** a total of 1,904 citation-form sign tokens.

## 3.3 The ASLLRP Sign Bank

The **ASLLRP Sign Bank**, available (via the **DAI 2**) from < **http://dai.cs.rutgers.edu/dai/s/signbank** >, was initially constructed to allow viewing of the **ASLLVD**. It has since been expanded to incorporate citation-form signs just mentioned in Section 3.2. It also now provides access to all of the continuous signing corpora just described in Section 2. Furthermore, the **Sign Bank** now also provides access to the continuous signing corpora, making it possible, for each entry, to view the corresponding segmented signs from the continuous signing corpora, as well as the complete utterances in which they occur.

The **Sign Bank** has a list at the left, arranged alphabetically by the glosses for the primary entries. Multiple "entry/variants" may be grouped under a single primary entry in cases where there are lexical variants or closely related forms. The number in parentheses after each entry/variant corresponds to the total number of examples of that sign in the **Sign Bank**.

One canonical example of each entry/variant is displayed on the right when the user clicks on that gloss. The right and left handshapes of the start and end of the sign are shown to the left and right of the thumbnails of the signer. In this case, since it is a one-handed sign, there is no handshape for the non-dominant hand. One can play that sign video by clicking on the button "Play Sign Video", as shown in **Figure 4**. Clicking on "Play Composite Video" will simultaneously play all of the examples of citation-form signs available from the **ASLLVD**, as is shown in **Figure 5**.

The lower part of the left-hand navigation area of **Figure 4** offers many options available for searches. The user can search by gloss text, 'related English words,' and/or phonological properties.

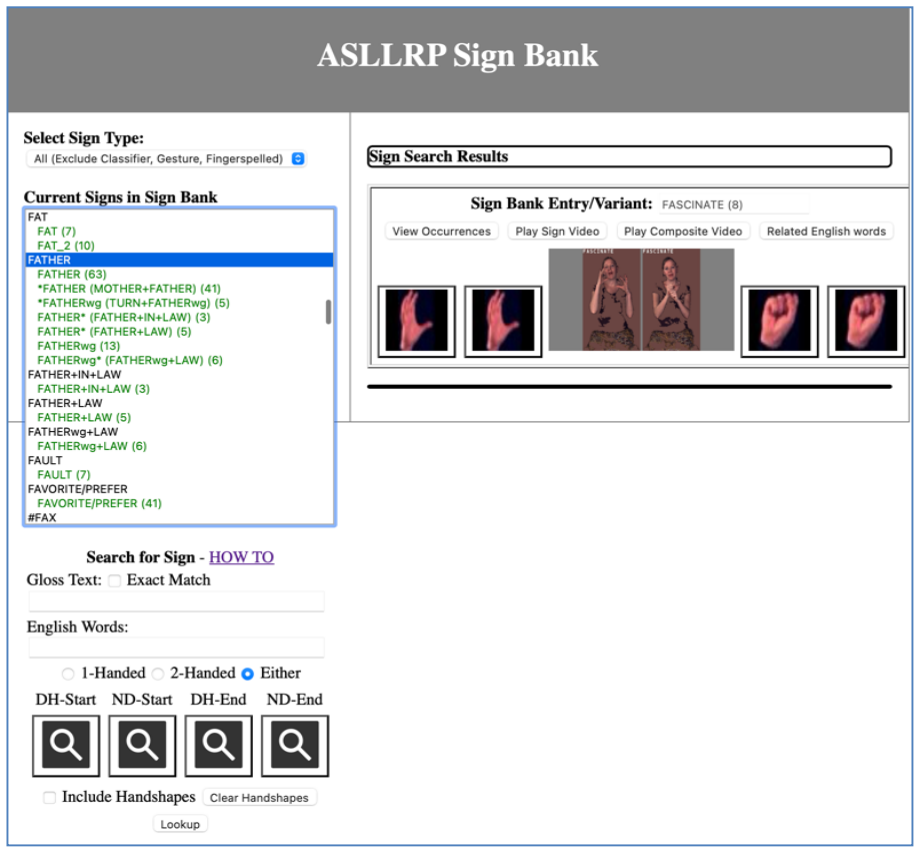

**Figure 4.** Alphabetical listing of *primary entries*, along with the corresponding *entry/variants*

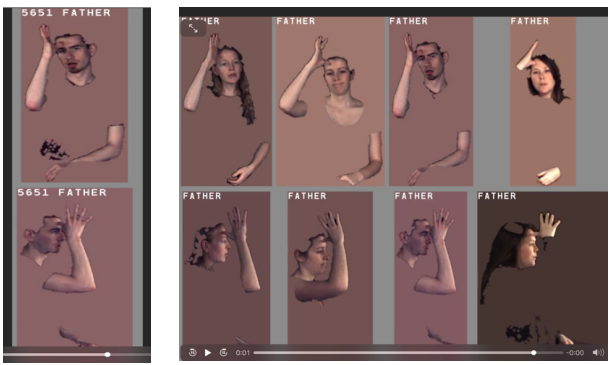   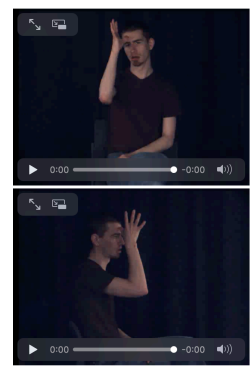

**Figure 5.** Playing the sign video (left) or the composite video (right)    **Figure 6.** Playing original video

Note that the **ASLLVD** video images on the left in **Figure 5** have undergone some post-processing.[1] There is also an option to "View Original Video, as shown in **Figure 6**."

---

[1] RAW frame captures → Bayer interpolation (un-calibrated color profile) → Skin color region detection and segmentation using skin color FG/BG model trained on signers in the data set → Background fill-in with average skin color → Crop the video frame to skin color segments → Color and contrast enhancement heuristic → Convert image sequence to .mov format video with gloss and frame number overlay. This was done was to facilitate presentation of video collated from several signers together for hand shape annotation purposes, to create videos that work within available scren real-estate and to provide consistent color, handshape contrast, and brightness across different signers.

Clicking "View Occurrences" will display examples in multiple parts of the window, arranged by data source: ASLLVD citation-form signs, segmented signs from the ASLLRP SignStream® 3 sentences, DawnSignPress (DSP) citation-form signs, segmented signs from DSP sentences, and RIT citation-form signs. On top, as seen in **Figure 7** are the examples of citation-form signs from the ASLLVD, and below are additional examples from the other data sets. For the continuous signing data, it is possible to view the video for just that sign, or for the entire utterance that contains the sign.

## 3.2 Navigation and Searching within the ASLLRP Sign Bank

It is now possible to limit the display to specific types of signs, from the top of the Sign Bank window shown in **Figure 4**; the options are shown in **Figure 8**.

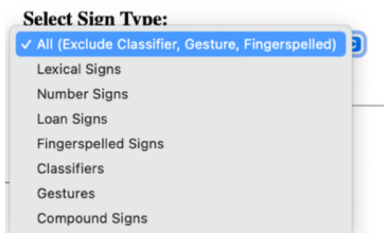

**Figure 8.** Selecting types of signs for Sign Bank display

If you click anywhere in the sign list shown in **Figure 4**, and then type a sequence of characters quickly, the selection will advance to the next sign that starts with those characters. So, if you are at the top of the list and start typing "FATH" the sign FATHER will be selected and will become visible. Alternatively, you can scroll to the entry you are interested in, or you can use the search tools at the bottom left in **Figure 4**. For more detailed information about how to use the DAI and about the options for searches, see [34] and https://www.bu.edu/asllrp/New-features-DAI2.pdf.

**Figure 7.** Display of results from multiple sources, for LIVE

### 3.4 Further Information

The conventions used for annotations in all of our corpora are documented in [29, 30]. A list of the field and value names and labels for non-manual events is also contained in the Appendix of [33] as is the structure of the XML export format. For additional information about **DAI 2** and the datasets to which it provides access, see [34-36]. Statistics about the contents of the various datasets are provided through **DAI 2** at < http://dai.cs.rutgers.edu/dai/s/runningstats >.

## 4. SignStream® – the Software used for the Linguistic Annotations

All of our video corpora have been linguistically annotated using **SignStream®**, an application for linguistic annotation of visual language data developed by our group and freely shared. It provides an intuitive interface for labeling and time-aligning manual and non-manual aspects of the signing while viewing up to 4 synchronized videos showing different views. See <http://www.bu.edu/asllrp/SignStream/3/>. The principal programmer is Gregory Dimitriadis, working with Douglas Motto at Rutgers University. A screen shot is shown in Figure 7. The annotations from **SignStream®** are uploaded to **DAI 2** to enable shared access, but are also available in an XML export format. See [31-33] for **SignStream® 3** User's Guides.

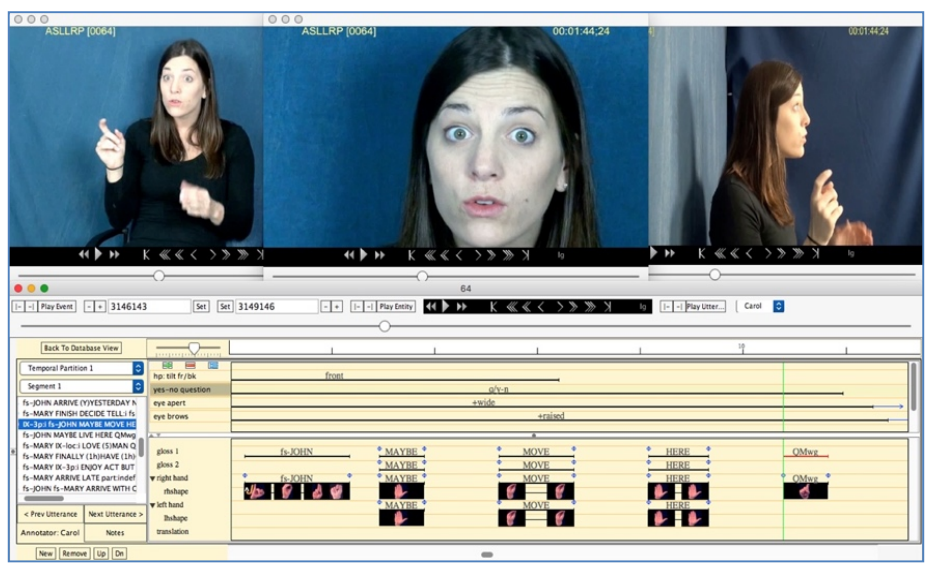

Figure 8. Screen shot of **SignStream**®

## 5. Research on Computer-Based ASL Recognition Enabled by These Data

The datasets just described—linguistically annotated video examples of native signers producing citation-form signs and continuous signing (utterances)—have been the basis for various types of linguistic research and for research on computer-based recognition of both manual signs and the non-manual components of ASL, by students and researchers all over the world , e.g. [1-6, 8-19, 22-25, 28, 39-45, 47-55, 57, 58, 60].

Our research group has been using these data for more than 20 years. In particular, we have done research on recognition of the grammatically significant non-manual expressions that co-occur with manual signing and of isolated, citation form signs. For detection of signs from continuous signing, we have done research to identify sequences containing specific types of signs (lexical, fingerspelled, and loan signs; classifiers; and gestures). Since these types have significantly different internal structures and are governed by differing linguistic constraints, distinct recognition strategies are needed. The following provides a brief overview of the types of research that have been enabled and facilitated by these data.

## 5.1 Recognition of Linguistically Significant Non-masnual Expressions

[26] "3D Face Tracking and Multi-scale, Spatio-temporal Analysis of Linguistically Significant Facial Expressions and Head Positions in ASL," Bo Liu, Jingjing Liu, Xiang Yu, Dimitris Metaxas, Carol Neidle (2012 – 8th International Conference on Language Resources and Evaluation (LREC). Istanbul, Turkey)

Essential grammatical information is conveyed in signed languages by clusters of events involving facial expressions and movements of the head and upper body. These can occur over hierarchical domains, co-occurring with the phrase over which they have scope. This poses a significant challenge for computer-based sign language recognition. The method we developed for the recognition of non-manual grammatical markers involved training based on our linguistically annotated corpora, which include annotations for both anatomical movements (e.g. head nods and shakes, eyebrow height, eye aperture) and the grammatical features they express (e.g., negation, question type, topic). Anatomical movements show characteristic patterns involving particular types of transitional movements; onsets and offsets are also annotated in our corpora. For example, periodic head movements typically include an initial rotation, so that the repeated head movement can begin with a maximum amplitude for the downward or sideward movement, with the periodic movement diminishing in amplitude over time. As illustrated in **Figure 9**, the side-to-side headshake that is one of the components of the non-manual marking of negation begins with a head rotation to the side, prior to the start of the grammatically significant portion of the negative marking, which extends over the scope of negation (e.g., the Verb Phrase), diminishing in amplitude over the course of articulation. Grammatical information expressed through changes in eyebrow height involves an anticipatory movement to get to the maximal or minimal height just before the beginning of the marking, as illustrated for the Topic in **Figure 9**.

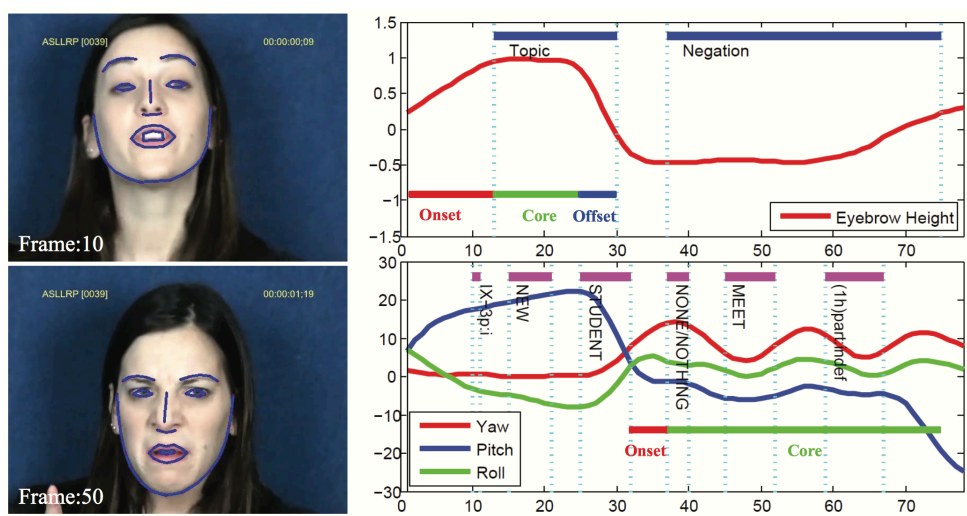

**Figure 9.** Eyebrow raise and head shake: "The new student, nobody has met (him/her)."

The method we developed used linguistically motivated multiscale features; the method was based on: (1) 3D deformable model-based tracking methods for the estimation of 3D head pose and facial expressions to determine the relevant low-level features; (2) methods for higher-level analysis of component events (raised/lowered eyebrows, periodic head nods and head shakes) used in grammatical markings—with differentiation of temporal phases (onset, core, offset, where appropriate), analysis of their characteristic properties, and extraction of corresponding features; and (3) a 2-level learning framework to combine low- and high-level features of differing spatio-temporal scales. This approach achieved significantly better tracking and recognition results than previous methods. **Figure 10** shows the methodology that combines low and high level features and non-manual events to recognize, using a 2-level conditional random field (CRF) learning-based method, non-manual grammatical markers in ASL.

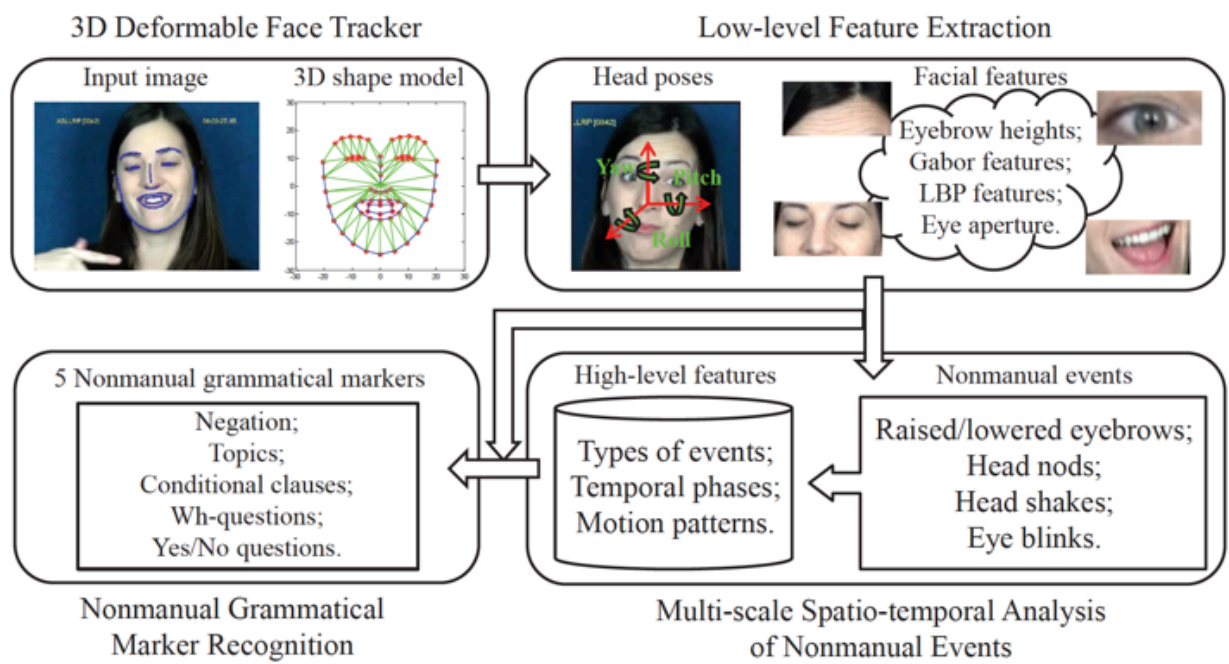

**Figure 10.** Flowchart of the approach

We used a 2-level CRF-based learning framework; non-manual markers and important temporal events are recognized and partitioned into the appropriate temporal phases. We evaluated non-manual event recognition using selected 85 videos from the **ASLLRP SignStream® 3 corpus**. These videos include different lighting conditions, partial hand occlusions, diverse head poses, and occasional motion blurrings. We successfully detected 79 raised eyebrow events of the 81 in the test dataset, 49/53 occurrences of lowered eyebrows, 38/42 head shake events, and 22/27 head nods. Finally, **Table 1** shows a confusion matrix of recognition of different non-manual markers using our approach.

|  | Wh-question | Negation | Topic | Yes/No question | Conditional/ when clause | None of the above |
|---|---|---|---|---|---|---|
| Wh-question | **6** | 0 | 0 | 0 | 0 | 1 |
| Negation | 0 | **34** | 0 | 0 | 0 | 1 |
| Topic | 0 | 0 | **46** | 0 | 6 | 3 |
| Yes/No question | 0 | 0 | 0 | **5** | 0 | 0 |
| Conditional/ when clause | 0 | 0 | 0 | 1 | **15** | 0 |
| None of the above | 1 | 0 | 1 | 1 | 0 | **0** |

**Table 1.** Confusion matrix for our face tracking system using both low- and high- level features. The label at the left of each row indicates the ground truth from the annotations.

## 5.2   Isolated Sign Recognition

[27] "Linguistically-driven Framework for Computationally Efficient and Scalable Sign Recognition," Dimitris Metaxas, Mark Dilsizian, Carol Neidle (2018 – 11th International Conference on Language Resources and Evaluation (LREC). Miyagawa, Japan)

We introduced a new general framework for isolated sign recognition from monocular video: a scalable, computational approach exploiting our linguistically annotated ASL datasets with multiple signers that uses both hand-crafted and deep neural network based features. We recognized signs using a hybrid framework combining learning methods with features based on what is known about the linguistic composition of lexical signs. We modeled and recognized the sub-components of sign production, with attention to hand shape, orientation, location, motion trajectories, plus non-manual features, and we combined these within a CRF framework; see **Figure 11**. This parameterization enables an extendable time-series learning approach, making the sign recognition problem robust, scalable, and feasible with relatively smaller datasets than are required for purely data-driven methods such as deep neural networks. From a 350-sign vocabulary of isolated, citation-form lexical signs from the **American Sign Language Lexicon Video Dataset (ASLLVD),** including both 1- and 2-handed signs, we achieved a top-1 accuracy of 93.3% and a top-5 accuracy of 97.9%. The high probability with which we could produce 5 sign candidates that contain the correct result opens the door to many potential applications, as it is reasonable to provide a sign lookup functionality that offers the user 5 possible signs, in decreasing order of likelihood, with the user then asked to select the desired sign.

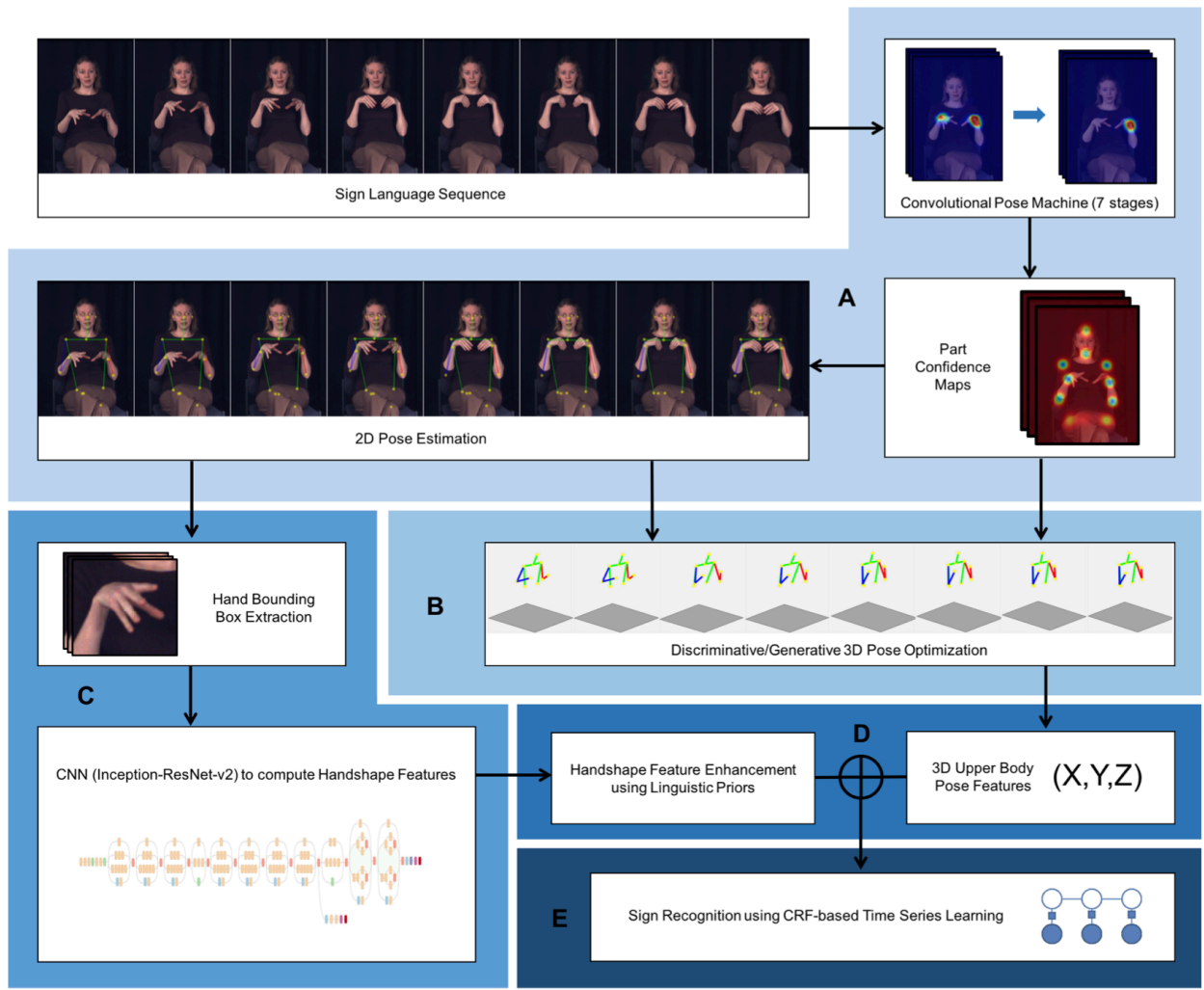

**Figure 11.** Overview of the approach, using handcrafted features extracted from skeleton data and the face, and, for the hands, features from a deep neural network.

More recently, we have developed a new deep-learning, skeleton-based method for isolated sign recognition that involves explicit detection of the start and end frames of signs, trained on the **ASLLVD** dataset; it uses linguistically relevant parameters based on the skeleton input. The new method employs a bidirectional learning approach within a Graph Convolutional Network (GCN) framework.[2] It achieves a success rate of 77.43% for top-1 and 94.54% for top-5 on a corpus (with gloss labels consistent with those used in the **ASLLRP Signtream® corpora**) that includes a vocabulary of about 1500 signs.

---

[2] "Bidirectional Skeleton-Based Isolated Sign Recognition using Graph Convolution Networks and Transfer Learning," Konstantinos M. Dafnis, Evgenia Chroni, Carol Neidle, Dimitris N. Metaxas *(in prep.)*

### 5.3 Detection of Major Sign Types in Continuous Signing for ASL Recognition

[59] "Detection of Major ASL Sign Types in Continuous Signing for ASL Recognition," Polina Yanovich, Carol Neidle, and Dimitris Metaxas (2016 – 10th International Conference on Language Resources and Evaluation (LREC). Portorož, Slovenia)

In ASL as in other signed languages, different classes of signs have different internal linguistically based structural properties, requiring distinct recognition strategies. For these strategies to be applied, continuous signing video needs to be segmented into strings of specific sign types. Our corpora include annotations of sign type, and have been used as training data for distinguishing strings of distinct sign types in continuous signing video, e.g. [56] which distinguished strings of fingerspelling within continuous signing.

For partitioning continuous signing into strings of specific sign types and identifying those sign types, we developed a multiple instance learning-based segmentation system. We formulated sign type classification as an intra-frame multi-instance (MIL) problem, allowing us to capture indirectly important relationships among multiple moving regions in the image. MIL captures local motion inside the frame, but does not capture global temporal changes. However, local motion within a frame can be consistent with multiple sign classes. Therefore, we used a one state per frame CRF [20] on top of the MIL framework output to model the global interframe dynamics. The system uses novel feature descriptors derived from both motion and shape statistics of the regions of high local motion. The system does not require a hand tracker. We accurately labeled 91.27% of the video frames of 500 continuous utterances (including 7 different subjects) from the **NCSLGR SignStream® 2 Corpus** [see Section 2.1].

### 5.4 Anonymization of ASL Videos

We have recently developed an optical flow based deep learning framework. Given a target image and a video sequence of continuous signing, we anonymize the original signing by mapping it to the person in the target image, preserving the linguistic information from the original signer. We have done preliminary user studies [21] that confirm the effectiveness of both the retention of linguistic information and the anonymization.

### 5.5 Computer-based Methods Using our Data

The above methodologies of incorporating linguistic modeling into machine learning frameworks are general and we are currently exploring new end-to-end deep learning architectures.

## 6. Conclusion

In sum, we have provided information about the datasets shared on our websites, for browsing, searching, and downloading. These datasets have proved invaluable in our own research, and we hope they may be of use to others for various types of research and educational applications.

*Acknowledgments: This research has been supported by grants from the National Science Foundation (see < http://www.bu.edu/asllrp/nsf.html >). Current work on this project is funded by NSF grants #1763523, 1763486, 1763569, and 2040638. See < **http://www.bu.edu/asllrp/people.html** > for a list of the many, many, many people who have made important contributions to the research that gave rise to the materials described here, including especially Gregory Dimitriadis, Douglas Motto (RU); Stan Sclaroff, Ashwin Thangali, Vassilis Athitsos, Dawn MacLaughlin, Robert Lee, and Joan Nash (BU); Ben Bahan and Christian Vogler (GU); and Matt Huenerfauth (RIT). The first version of the DAI was implemented primarily by Christian Vogler (following up on preliminary work in which many people had been involved) [38]. Subsequent development has been carried out by Augustine Opoku, who has designed and developed DAI 2, with many additional features beyond those available in the original DAI.

**Data Availability:** The websites through which the resources are accessible, at no cost, subject to stated terms of use, include:

- **ASLLRP Sign Bank**   **https://dai.cs.rutgers.edu/dai/s/signbank** - for viewing **ASLLVD** and segmented signs and their utterance contexts from **ASLLRP continuous signing corpora**):
- **DAI 2**   **https://dai.cs.rutgers.edu/dai/s/dai** - **for browsing, searching, and downloading continuous signing data from:**
    - **ASLLRP SignStream® 3 Corpus**: https://dai.cs.rutgers.edu/dai/s/dai
    - **NSCLGR SignStream® 2 Corpus**: https://dai.cs.rutgers.edu/dai/s/daioriginal
- **Annotation software**, **SignStream® 3:** http://www.bu.edu/asllrp/SignStream/3/

# Table of Contents